\documentclass[journal]{IEEEtran}

\usepackage{bbding}
\usepackage{graphicx}
\usepackage{amssymb}
\usepackage{wrapfig} 
\usepackage[linesnumbered, ruled]{algorithm2e}
\usepackage{url}
\usepackage{booktabs}
\usepackage{multirow}
\usepackage{amsmath}

\usepackage[table]{xcolor}

\usepackage[pagebackref=true,breaklinks=true,colorlinks,bookmarks=false,citecolor=blue]{hyperref}

\definecolor{lightblue}{RGB}{80,160,220}

\ifCLASSINFOpdf
\else
\fi
\begin{document}
%

\title{Consistent Evidence, Robust Recognition: Faithful Attribution Regularization under Geometric Transformations}
%
%
%

\author{Xianghao~Jiao,
        Ruoyu~Chen,
        Wei~Wang,
        Jiazi~Hu,
        Jiawei~Liang,
        Shangquan Sun,
        Shiming~Liu,\\
        Qunli~Zhang,
        and~Xiaochun~Cao,~\IEEEmembership{Senior~Member,~IEEE}
\thanks{Xianghao Jiao, Wei Wang, Jiawei Liang, and Xiaochun Cao are with the School of Cyber Science and Technology, Shenzhen Campus of Sun Yat-sen University, Shenzhen, China (Email: \href{mailto:jiaoxh0331@outlook.com}{jiaoxh0331@outlook.com}, \href{mailto:wangwei29@mail.sysu.edu.cn}{wangwei29@mail.sysu.edu.cn}, \href{mailto:liangjw57@mail2.sysu.edu.cn}{liangjw57@mail2.sysu.edu.cn}, \href{mailto:caoxiaochun@mail.sysu.edu.cn}{caoxiaochun@mail.sysu.edu.cn}).}
\thanks{Ruoyu Chen is with the University of Chinese Academy of Sciences, Beijing, China 
(E-mail: \href{mailto:cryexplorer@gmail.com}{cryexplorer@gmail.com}).}
\thanks{Jiazi Hu is with the School of Automation, Nanjing University of Information Science and Technology (Email: \href{mailto:hujiazi@nuist.edu.cn}{hujiazi@nuist.edu.cn}).}
\thanks{Shiming Liu and Qunli Zhang are with the Department of Mechanical Engineering, Imperial College London, UK (E-mail: \href{mailto:852074479@qq.com}{852074479@qq.com}, \href{mailto:kingsleyzhang@qq.com}{kingsleyzhang@qq.com}).}
\thanks{Shangquan Sun is with the College of Computing and Data Science, Nanyang Technological University, Singapore (Email: \href{mailto:shangquan.sun@ntu.edu.sg}{shangquan.sun@ntu.edu.sg}).}
\thanks{Xianghao Jiao and Ruoyu Chen contributed equally to this work.}
\thanks{Corresponding author: Xiaochun Cao.}
}

%



\maketitle

\begin{abstract}
    Attribution methods are widely used to characterize the evidence underlying model predictions, yet their potential to improve model behavior remains underexplored. Attribution inconsistency under label-preserving geometric transformations may indicate transformation-sensitive evidence reliance, motivating attribution regularization. However, such supervision is valid only when attribution faithfully reflects the evidence driving predictions. Existing self-supervised methods typically align gradient-based maps such as Grad-CAM, whose limited faithfulness means that attribution consistency need not imply consistency of the underlying decision process, leaving transformation robustness unresolved. We propose an annotation-free attribution regularization framework based on submodular search over image regions. By measuring how candidate subsets affect model outputs, the search extracts compact, class-discriminative evidence as search-derived supervision. We further introduce a submodular ranking loss with path-consistency and termination-alignment terms that respectively align spatially corresponding candidate rankings along paired search trajectories and encourage the transformed trajectory to satisfy the stopping criterion at the target terminal step. The loss provides a differentiable surrogate for regularizing both final attributions and the otherwise discrete evidence-selection process. Experiments on ImageNet-100 show that our method raises attribution stability from 0.14 to 0.27, improves Insertion by 52.1\%, and reduces Deletion by 62.2\% on ViT-B/16, with only a 0.28-percentage-point accuracy drop; consistent gains are observed on ViT-L/16. On ImageNet-1K, it improves mean transformed-input accuracy by 0.55 and 0.33 percentage points on ResNet-50 and ConvNeXt-B, respectively, while limiting the clean-accuracy drop to 0.30 points. These results show that faithful attribution promotes consistent evidence reliance without materially compromising predictive performance. The code will be released soon.
\end{abstract}


%
\IEEEpeerreviewmaketitle

\newpage
\section{Introduction}

\begin{figure*}[t]
\centering
\includegraphics[width=1.0\textwidth]{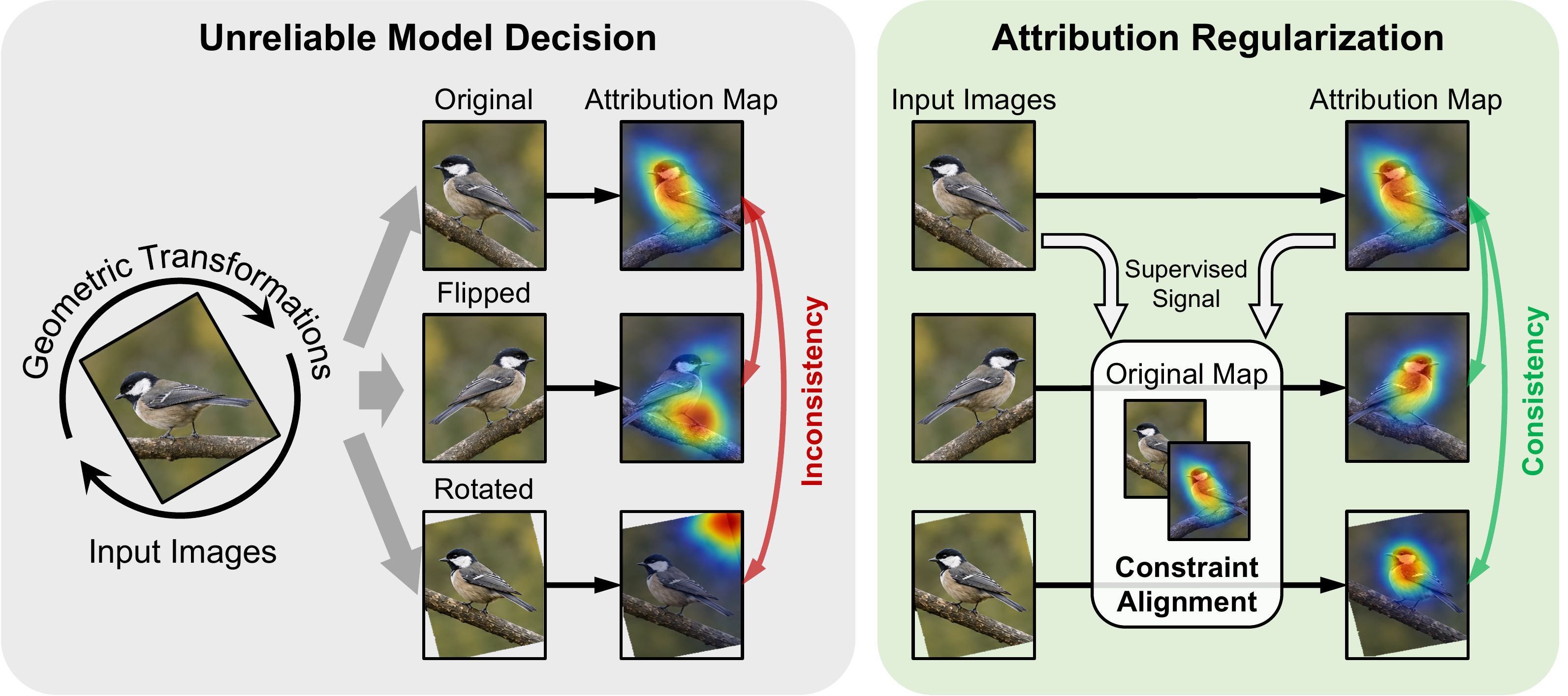}
\caption{Motivation and overview of the proposed attribution-regularization framework. Under label-preserving geometric transformations, a model may shift its attribution to inconsistent regions, revealing unreliable decision evidence (left). Our framework uses the attribution of the original image as a supervision signal and aligns the attributions of transformed inputs with their spatially corresponding targets, encouraging consistent evidence reliance across transformations (right).}
\label{fig:first_figure}
\end{figure*}

\IEEEPARstart{F}{eature} attribution methods~\cite{chen2026mllms,chen2025less,IG17,IG2-24,path24,eclipC24,lundberg2017unified} are widely used to identify the input evidence associated with individual predictions of deep neural networks. Most attribution research treats explanations as post-hoc diagnostic outputs. A complementary opportunity is to use the evidence revealed by attribution to improve the model itself~\cite{xaisurvey24,chen2025generalized,chen2026not}. Consider a label-preserving geometric transformation applied to an image. Even when the predicted class remains unchanged, the model may shift from the original object evidence to a different or irrelevant region~\cite{d2022underspecification,geirhos2018imagenet}. A faithful attribution should expose this change; conversely, if the model relies on spatially corresponding evidence, its attribution should transform equivariantly with the input~\cite{invariacemetric23}. As illustrated in the left panel of Fig.~\ref{fig:first_figure}, flipping or rotating an image can redirect the model's attribution to inconsistent regions despite preserving its semantic content. The pronounced inconsistency observed for many explanation methods under small input changes~\cite{explain-robustness18} therefore motivates attribution regularization as a means of encouraging consistent evidence reliance.

The validity of this training signal depends critically on attribution faithfulness: an explanation must reflect the evidence that actually affects the model's prediction~\cite{chen2025less,kim2022hive}. Existing explanation-guided learning methods either align attributions with human-provided relevance annotations~\cite{schramowski2020making,rrr17,hint19,amc23} or construct annotation-free consistency objectives from paired inputs~\cite{gc21,cgc22,dre23,LiuWY23}. The latter are particularly attractive because geometric transformations provide spatial correspondence without additional labeling. In practice, however, these methods commonly regularize unfaithful saliency maps, such as Grad-CAM~\cite{gradcam20}. Although computationally convenient, the faithfulness and parameter sensitivity of such maps can be limited~\cite{chen2025less,sanity18,KarimiMKSK23}. Consequently, agreement between two attribution maps does not necessarily imply that the underlying prediction process relies on corresponding evidence. A model may therefore achieve explanation-level consistency while remaining sensitive to geometric transformations. This distinction leads to the central question of this work: \emph{\textbf{how can faithful, decision-linked attribution be converted into an effective training signal for consistent evidence reliance?}}

As illustrated in the right panel of Fig.~\ref{fig:first_figure}, we address this question with an annotation-free attribution-regularization framework based on submodular search over image regions~\cite{lima24}. Our framework uses the attribution of the original image as a supervision signal and constrains transformed inputs to rely on spatially corresponding evidence. Specifically, rather than deriving a heatmap from local gradients, the search explicitly evaluates how candidate region subsets affect the model output and greedily constructs an ordered sequence of compact, class-discriminative evidence. In the query stage, we retain correctly classified, high-confidence samples whose resulting attribution occupies a sufficiently small area. For each retained sample, the selected sequence and its stopping condition constitute search-derived supervision. A geometric transformation then maps this sequence to spatially corresponding target regions for the transformed image. This filtering and query procedure ties the supervision to model behavior while reducing the influence of diffuse or unreliable search results.

In the training stage, optimizing this supervision is non-trivial because search-based attribution is discrete, sequential, and path-dependent. Matching only the final selected set neither differentiates through the greedy choices nor ensures that the transformed input reaches the result through a corresponding evidence-selection process. We therefore introduce a \emph{Submodular Ranking Loss} with two complementary components. At each search step, the selection-ranking term enforces path consistency by ranking the spatially corresponding target region above all remaining candidates under the transformed input. At the final target step, the selection-truncation term encourages the accumulated transformed evidence to satisfy the same stopping criterion as the query sequence. Together, the two terms provide a differentiable surrogate for regularizing both the final attribution and the otherwise discrete evidence-selection trajectory. The resulting two-stage framework combines the \emph{Submodular Ranking Loss} with the classification objective to update the model while preserving its predictive capability.

Experiments support the connection between faithful attribution supervision and consistent evidence reliance. On ImageNet-100 with ViT-B/16, our method increases attribution stability from 0.14 to 0.27, improves Insertion by 52.1\%, and reduces Deletion by 62.2\%, while decreasing classification accuracy by only 0.28 percentage points; consistent improvements are observed with ViT-L/16. On ImageNet-1K, the method improves mean transformed-input accuracy by 0.55 and 0.33 percentage points for ResNet-50 and ConvNeXt-B, respectively, while limiting the clean-accuracy reduction to at most 0.30 points. These results indicate that the proposed method improves not only agreement between attribution outputs but also the model's reliance on corresponding decision evidence under geometric transformations.

The main contributions are summarized as follows:
\begin{itemize}
    \item We formulate attribution regularization around a prerequisite not guaranteed by map-consistency objectives: the attribution used for supervision must faithfully represent the evidence underlying the model prediction. Based on this principle, we develop an annotation-free framework that derives compact supervision sequences from filtered submodular searches.
    \item We propose a Submodular Ranking Loss that aligns candidate selection along the transformed trajectory with the query sequence and encourages the terminal subset to meet the corresponding stopping criterion, providing a differentiable surrogate for a discrete, path-dependent attribution process.
    \item We conduct extensive experiments across ImageNet-100 and ImageNet-1K, multiple network families, and complementary attribution and prediction metrics, demonstrating improved attribution stability and faithfulness, stronger transformed-input accuracy, and only minor changes in clean-image performance.
\end{itemize}

\section{Related Work}
\subsection{Search-based Attribution Methods}

Search-based attribution methods generate high-precision attribution maps by searching the input space for the local regions most critical to a model's prediction and scoring candidate regions with an evaluation function.
For example, SAGs~\cite{shitole2021one} employs a beam search algorithm to systematically explore multiple high-confidence local regions within an input image, thereby moving beyond the limitation of traditional saliency maps, which can only generate a single explanation.
MoXI~\cite{SumiyasuKK24}, a game theory-based attribution method, leverages Shapley values to identify critical pixel groups by quantifying their synergistic interactions, rather than evaluating each pixel's contribution in isolation, thereby revealing the basis for model decisions with greater accuracy.
LIMA~\cite{lima24,chen2025less} is an attribution method motivated by submodular optimization. It employs a search strategy to filter the most representative key regions from sparsified image patches according to multiple scoring criteria, thereby generating concise and interpretable attribution maps.
VPS~\cite{vps25} is a novel attribution method based on visual precision search, specifically designed to interpret multimodal foundation models (e.g., Grounding DINO~\cite{LiuZRLZYJLYSZZ24} and Florence-2~\cite{0004WXDHL00Y24}). By partitioning input images into sparse sub-regions and leveraging submodular functions for region selection, it generates highly precise attribution maps.
These methods use search primarily to generate post-hoc explanations. In contrast, our method turns LIMA's ordered selection path and stopping condition into self-generated training targets, and uses them to regularize the predictor's evidence selection under geometric transformations.

\subsection{Explanation Robustness and Equivariance}

Explanation robustness requires explanations to change predictably under small or semantics-preserving changes to the input. Alvarez-Melis and Jaakkola~\cite{alvarez2018robustness} introduced a local Lipschitz-based criterion for quantifying this property and showed that nearby inputs can receive substantially different explanations. Ghorbani et al.~\cite{ghorbani2019interpretation} further demonstrated that label-preserving, visually imperceptible perturbations can drastically alter feature-attribution maps. Moving from generic perturbations to structured transformations, Crabb\'e and van der Schaar~\cite{crabbe2023evaluating} formalized symmetry-respecting explanations through invariance and equivariance, and derived quantitative metrics and theoretical guarantees for evaluating interpretability methods. This line of work primarily diagnoses explanation fragility or makes the post-hoc explanation procedure more symmetry-aware while treating the predictor as fixed. Consequently, it can reveal or reduce inconsistency in explanation outputs, but does not rectify the transformation-sensitive evidence reliance learned by the underlying model. In contrast, we use attribution equivariance as a self-supervised training signal and update the predictor to select spatially corresponding, decision-relevant evidence across transformed inputs.


\subsection{Attribution Regularization}
Attribution regularization~\cite{xaisurvey24} enhances model training by applying constraints directly to attribution maps, guiding the model to minimize reliance on noise and irrelevant features. This encourages the model to focus on meaningful patterns in the data, significantly improving the rationality and reliability of attribution results.
Consistency is a core objective of attribution regularization, ensuring that the model's attribution results remain stable even when the input data distribution undergoes significant changes.
Several methods have now been proposed to enhance the consistency of attribution regularization. For example, Pillai et al.~\cite{gc21} proposed improving the consistency of model explanations by incorporating consistency constraints during training, resulting in more coherent and reliable explanations under a given attribution method. Li et al. proposed the DRE framework~\cite{dre23}, which leverages the differences between data distributions as a self-supervised signal to guide the model toward producing consistent and stable explanations for the same input, even under distribution shifts or environmental variations. This approach significantly enhances the robustness and reliability of model attributions in out-of-distribution (OOD) scenarios. Liu et al. proposed the ICEL~\cite{LiuWY23} method, which introduces an Inconsistent Explanation Loss as its core component. This loss function optimizes the model by measuring the differences between explanation heatmaps corresponding to different predictions for the same image. These studies demonstrate the value of explanation-guided training, but the closest transformation-consistency methods, such as GC~\cite{gc21} and CGC~\cite{cgc22}, construct their supervision from local gradient- or activation-based maps. Our method instead derives high-faithfulness supervision from output responses measured during submodular region search and introduces a differentiable ranking objective for the otherwise discrete search process, thereby regularizing the model without human attribution annotations.

\begin{figure*}[t]
    \centering
    \includegraphics[width=1.0\textwidth]{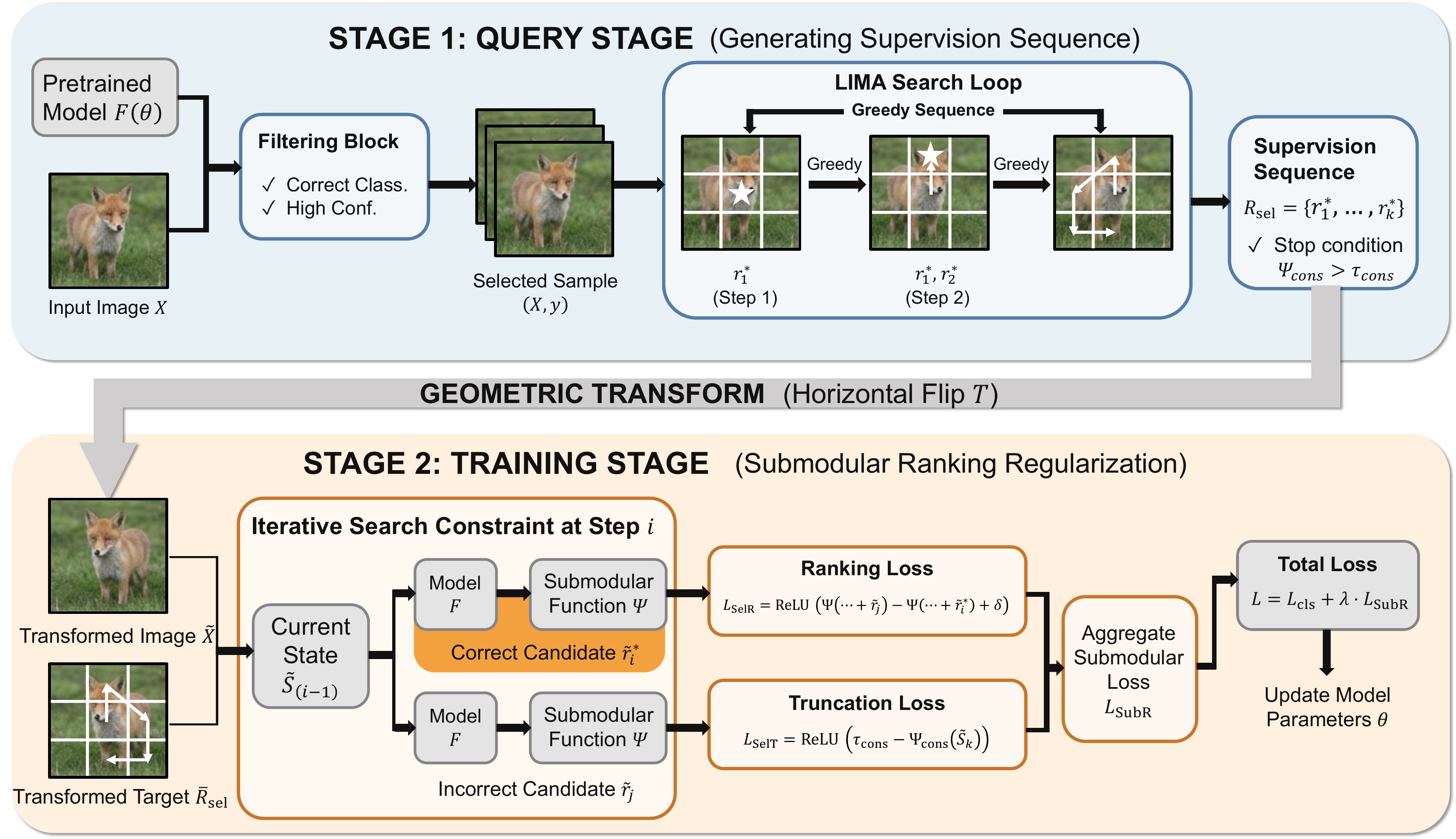}  
    \caption{Overview of the proposed attribution-regularization framework and its two-stage computation. In the query stage, the original image is searched to obtain an ordered decision-region sequence $R_{\text{sel}}$, and the search stops when the consistency score reaches the predefined threshold $\tau_{\text{cons}}$. This sequence serves as the supervision target. During training, the selection-ranking loss $\mathcal{L}_{\text{SelR}}$ encourages each target region to rank above the remaining candidates, while the selection-truncation loss $\mathcal{L}_{\text{SelT}}$ encourages the target terminal subset to satisfy the stopping criterion.}
    \label{fig:main_figure}
\end{figure*}

\section{Method}

This section presents our attribution regularization method, as shown in Fig.~\ref{fig:main_figure}. Sec.~\ref{problem_formulation} formulates the objective of transformation-consistent regularization, and Sec.~\ref{sec:search_attribution} introduces search-based attribution for generating faithful, ordered supervision. Sec.~\ref{sec:submodular_ranking_loss} presents the Submodular Ranking Loss, which aligns region selection across transformed search trajectories and enforces the stopping criterion on the terminal subset. Sec.~\ref{sec:search_regularization} explains how this loss is used to fine-tune a pretrained model, including the optimization objective, reliability filtering, and two-stage procedure. Finally, Sec.~\ref{sec:training_efficiency} discusses the implementation choices used to improve training efficiency.

\subsection{Problem Formulation}\label{problem_formulation}

Let $\mathcal{D}$ denote a training distribution of image--label pairs $(X,y)$, and let $\mathcal{F}_{\theta}$ be a predictive model parameterized by $\theta$. We consider a label-preserving geometric transformation $t$ sampled from a transformation distribution $\mathcal{T}$, yielding a transformed view $\tilde{X}=t(X)$. For brevity, let $\mathcal{A}_{\theta}(X,y)$ denote the attribution produced by $\mathcal{F}_{\theta}$ for class $y$. Because an attribution is spatially tied to its input, the desired transformation invariance is more precisely an \emph{equivariance} condition: the attribution of the transformed image should coincide with the spatially transformed attribution of the original image,
\begin{equation}
    \mathcal{A}_{\theta}(t(X),y)
    \approx
    t\!\left(\mathcal{A}_{\theta}(X,y)\right).
\end{equation}

Accordingly, transformation-consistent attribution regularization can be formulated as
\begin{equation}
\label{eq:problem_formulation}
 \begin{aligned}
    \min_{\theta}\quad&
    \mathbb{E}_{\substack{(X,y)\sim\mathcal{D}\\t\sim\mathcal{T}}}
    \Big[
    \underbrace{\mathcal{L}_{\mathrm{task}}\!\left(\mathcal{F}_{\theta}(X),y\right)}_{\text{task supervision}}
    \\
    &\quad+
    \underbrace{\lambda\,
    \mathcal{L}_{\mathrm{attr}}\!\left(
        \mathcal{A}_{\theta}(t(X),y),
        t\!\left(\mathcal{A}_{\theta}(X,y)\right)
    \right)}_{\text{attribution regularization}}
    \Big],
 \end{aligned}
\end{equation}
where $\mathcal{L}_{\mathrm{attr}}$ is a single regularization term measuring the inconsistency between spatially corresponding attributions, and $\lambda$ controls its strength. This objective encourages the model to rely on corresponding decision evidence before and after transformation while retaining predictive performance. For a search-based attribution, however, $\mathcal{A}_{\theta}$ is a discrete, sequential procedure and cannot be optimized directly. We therefore instantiate $\mathcal{L}_{\mathrm{attr}}$ with the differentiable Submodular Ranking Loss introduced below, which aligns both the search trajectory and its termination.

\subsection{Preliminary: Search-based Attribution}\label{sec:search_attribution}

Attribution regularization is effective only when the attribution used as supervision faithfully reflects the evidence underlying the model's prediction. We therefore adopt search-based subset attribution for its high faithfulness~\cite{lima24,chen2025less}. Rather than relying on local gradients, this class of methods directly probes a model using different retained or masked region subsets, thereby tying the resulting explanation to changes in the model output. Prior studies, such as LIMA~\cite{lima24,chen2025less} and EAGLE~\cite{chen2026mllms}, have empirically shown that submodular subset-selection methods can identify compact, class-discriminative evidence with high faithfulness across diverse model architectures~\cite{lima24,chen2025less}. This decision-linked property makes them particularly suitable for constructing attribution supervision. We next briefly introduce the attribution algorithm used in this work; further details are provided in \textit{supplementary material}~\ref{append:submodular}.

Given an image partitioned into candidate regions, the attribution process greedily searches for a subset $S$ that maximizes an interpretability score $\Psi(S,\mathcal{F})$. This procedure is inspired by submodular maximization, for which greedy selection admits a standard approximation guarantee when the objective is monotone and submodular~\cite{NemhauserWF78}. The score used in this work consists of three components:

\textbf{Consistency score}. The consistency score measures how strongly the selected region set $S_k$ supports the prediction for the full image and is defined as
\begin{equation}
    \Psi_{\text{cons}}(S_k, \mathcal{F})
    =\bigl[\mathcal{F}(S_k)\bigr]_{\bar{y}},
\end{equation}
where $\mathcal{F}$ represents the classification model to be explained, $\bar{y}$ is the predicted category, and the selected $S_k$ is derived by merging the currently selected regions $R_{\text{sel}}=\{r_1^*,\ldots,r_k^*\}$, denoted as $S_k=\operatorname{SUM}(R_{\text{sel}})$.

\textbf{Confidence score}. The confidence score measures the predictive confidence obtained from the selected region set $S_k$ and is defined as
\begin{equation}
    \Psi_{\text{conf}}(S_k, \mathcal{F})
    =1-\frac{C}{\sum_{i=1}^C
    \left(\exp\!\left(\bigl[\mathcal{F}(S_k)\bigr]_i\right)+1\right)}.
\end{equation}

\textbf{Collaboration score}. The collaboration score measures how little evidence for the prediction remains after the selected region set $S_k$ is removed and is defined as
\begin{equation}
    \Psi_{\text{colla}}(S_k, X, \mathcal{F})
    =1-\Psi_{\text{cons}}(X-S_k, \mathcal{F}).
\end{equation}

The overall submodular function is calculated as:
\begin{equation}
\label{eq:subf}
    \Psi=\lambda_1\cdot\Psi_{\text{cons}} + \lambda_2\cdot\Psi_{\text{conf}} + \lambda_3\cdot\Psi_{\text{colla}}.
\end{equation}

In practice, because $\Psi$ is computed from the nonlinear responses of the model being explained, strict submodularity is not assumed to hold for every model and input. We therefore regard this procedure as a submodular-optimization-inspired greedy attribution algorithm rather than relying on a formal approximation guarantee. Nevertheless, its strong empirical attribution performance makes it an effective mechanism for extracting decision-relevant supervision.

\subsection{Submodular Ranking Loss}\label{sec:submodular_ranking_loss}

Conventional attribution methods generate results in one step, whereas search-based methods produce explanations through a sequential decision process. Therefore, enforcing attribution equivariance under geometric transformations requires the greedy search paths of the original and transformed images to select spatially corresponding regions.
Motivated by this observation, we use the submodular score $\Psi(S,\mathcal{F}_{\theta})$, which depends on the model parameters $\theta$ and measures how well a selected region set supports the prediction. Based on the greedy search trajectory, we propose a novel \textit{Submodular Ranking Loss} that constrains the order of region selection to be equivariant to geometric transformations. The overall pipeline is illustrated in Fig.~\ref{fig:main_figure}.

Given a training image--label pair $(X,y)$, we first obtain an attribution-region query sequence for the image. The image is partitioned into $n$ subregions to form the candidate set $R_{\text{can}}=\{r_1,\ldots,r_n\}$. We then employ LIMA to search for attribution regions. The process terminates when the consistency score $\Psi_{\text{cons}}(S_k,\mathcal{F})$ of the currently selected subset reaches a predefined threshold $\tau_{\text{cons}}$ (set to $\tau_{\text{cons}}=0.7$ in the experiments). The merged region $S_k$ is taken as the final attribution, and the corresponding ordered sequence $R_{\text{sel}}=\{r_1^*,\ldots,r_k^*\}$ is used as the supervision target during training.

During training, we require the attribution-search sequence of a transformed image to follow the transformed query sequence. Using horizontal flipping as the geometric transformation, we obtain the flipped candidate set $\tilde{R}_{\text{can}}=\{\tilde{r}_1,\ldots,\tilde{r}_n\}$ and target sequence $\tilde{R}_{\text{sel}}=\{\tilde{r}_1^*,\ldots,\tilde{r}_k^*\}$. At step $i$, let $\tilde{R}_{i-1}=\{\tilde{r}_1^*,\ldots,\tilde{r}_{i-1}^*\}$ denote the regions already selected and let $\tilde{\mathcal{C}}_i=\tilde{R}_{\text{can}}\setminus(\tilde{R}_{i-1}\cup\{\tilde{r}_i^*\})$ contain the remaining non-target candidates. We define $q_i(\tilde r)=\Psi(\operatorname{SUM}(\tilde{R}_{i-1}\cup\{\tilde r\}),\mathcal{F})$. The target $\tilde r_i^*$ should score higher than every candidate in $\tilde{\mathcal{C}}_i$. Because only the top-ranked region matters, rather than a strict ordering of all candidates, we define the \textit{selection-ranking loss} using a pairwise margin objective:
\begin{equation}
    \mathcal{L}_{\texttt{SelR}}
=\frac{
\sum\limits_{i=1}^{k}
\sum\limits_{\tilde r\in\tilde{\mathcal{C}}_i}
\operatorname{ReLU}\!\left(
q_i(\tilde r)-q_i(\tilde r_i^*)+\delta
\right)}
{\sum\limits_{i=1}^{k}\lvert\tilde{\mathcal{C}}_i\rvert},
\end{equation}
where $\delta$ controls the score margin between the target region $\tilde{r}^*_i$ and each remaining candidate. At $i=1$, $\tilde{R}_0=\varnothing$, and $\tilde{S}_0=\operatorname{SUM}(\tilde{R}_0)$ is the all-zero starting image. In addition to path consistency, the transformed trajectory should satisfy the stopping criterion at the target terminal step. Let $\tilde{S}_k=\operatorname{SUM}(\tilde{R}_{k-1}\cup\{\tilde r_k^*\})$. We define the \textit{selection-truncation loss} as
\begin{equation}
    \mathcal{L}_{\texttt{SelT}}
    =\operatorname{ReLU}\!\left(
    \tau_{\text{cons}}
    -\Psi_{\text{cons}}(\tilde{S}_k,\mathcal{F})
    +\delta
    \right).
\end{equation}


The resulting \textit{Submodular Ranking Loss} is
\begin{equation}
    \label{eq:subr}
    \mathcal{L}_{\texttt{SubR}}
    =\mathcal{L}_{\texttt{SelR}}+\mathcal{L}_{\texttt{SelT}}.
\end{equation}

\begin{algorithm}[t]
  \KwIn{Pretrained model $\mathcal{F}$ with parameters $\theta$, training set $\mathcal{D}$, transformation distribution $\mathcal{T}$, search-termination threshold $\tau_{\text{cons}}$, regularization strength $\lambda$, classification confidence threshold $\tau_{\text{conf}}$, attribution-area threshold $\tau_a$, and learning rate $\alpha$.}
  \KwOut{Fine-tuned model $\mathcal{F}^*$.}
  \For{$e=1,\ldots,N$}{
    \For{training sample $(X, y)\in\mathcal{D}$}{
        Obtain the model prediction $z\leftarrow\mathcal{F}_{\theta}(X)$

        Divide image $X$ into region set $R=\{r_1,\ldots,r_n\}$
        
        Obtain the attribution-region sequence $R_{\text{sel}}=\{r_1^*,\ldots,r_k^*\}\leftarrow\operatorname{LIMA}(\mathcal{F},R,\tau_{\text{cons}})$
        
        Compute the classification loss $\mathcal{L}_{\text{cls}}\leftarrow\operatorname{CE}(z,y)$;
        
        \eIf{$\operatorname*{arg\,max}_{c}z_c=y$ and $\max_c z_c>\tau_{\text{conf}}$ and $\operatorname{Area}(S_k)<\tau_a$}{
            \tcp{Filter reliable sample}
            Sample a geometric transformation $t\sim\mathcal{T}$
            $(\tilde{X},\tilde{R}_{\text{sel}})\leftarrow\bigl(t(X),t(R_{\text{sel}})\bigr)$

            Compute the regularization using Eq.~\ref{eq:subr}: $\mathcal{L}_{\text{reg}}\leftarrow\mathcal{L}_{\text{SubR}}(\mathcal{F},\tilde{X},\tilde{R}_{\text{sel}})$;
            
            $\mathcal{L}\leftarrow \mathcal{L}_{\text{cls}}+\lambda \cdot \mathcal{L}_{\text{reg}}$;
        }{
            \tcp{Use only task supervision for an unreliable sample}
            $\mathcal{L}\leftarrow \mathcal{L}_{\text{cls}}$;
        }
        Update model parameters $\theta \leftarrow \theta-\alpha\nabla_{\theta}\mathcal{L}$;
    }
  }
        
  \caption{Attribution Regularization}\label{algo}
\end{algorithm}

\subsection{Search-based Attribution Regularization}\label{sec:search_regularization}

Having defined the Submodular Ranking Loss, we use it as a differentiable regularizer to fine-tune a pretrained model. Starting from a pretrained model is necessary because attribution search requires meaningful predictive behavior from which decision-relevant regions can be queried. Moreover, because the query sequence $R_{\text{sel}}$ is generated without attribution annotations, its reliability directly affects the fine-tuning outcome.

We optimize the model parameters using the following objective:
\begin{equation}
\label{eq:overall_objective}
\mathcal{L}
=\mathcal{L}_{\text{cls}}
+\lambda\,g(X,y)\mathcal{L}_{\texttt{SubR}},
\end{equation}
where $\mathcal{L}_{\text{cls}}$ preserves task performance, $\mathcal{L}_{\texttt{SubR}}$ regularizes the transformed attribution-search trajectory, and $\lambda$ balances the two terms. The indicator $g(X,y)$ determines whether the query sequence is sufficiently reliable for attribution regularization:
\begin{equation}
\begin{aligned}
g(X,y)=\mathbf{1}\big[
&\operatorname*{arg\,max}_{c}\,[\mathcal{F}_{\theta}(X)]_c=y,\\
&\max_c[\mathcal{F}_{\theta}(X)]_c>\tau_{\text{conf}},\
\operatorname{Area}(S_k)<\tau_a
\big].
\end{aligned}
\end{equation}

Regularization is thus applied only to correctly classified, high-confidence samples whose terminal attribution occupies less than the area threshold $\tau_a$. The first two conditions avoid reinforcing erroneous evidence from uncertain or incorrect predictions. The area condition excludes diffuse LIMA attributions, which may contain substantial irrelevant content and therefore provide weak supervision. If a sample fails any condition, $g(X,y)=0$, and it contributes only the classification loss.

Fine-tuning proceeds in two stages for each sample:

\textbf{Query stage.} We partition the original image into candidate regions and run LIMA on the current model until $\Psi_{\text{cons}}$ reaches $\tau_{\text{cons}}$, producing the ordered target sequence $R_{\text{sel}}$ and its terminal subset $S_k$. We then evaluate $g(X,y)$ to decide whether this query is used for regularization.

\textbf{Training stage.} For a reliable query, we sample $t\sim\mathcal{T}$ and spatially transform both the image and the queried sequence. The transformed sequence is treated as the supervision target: $\mathcal{L}_{\texttt{SelR}}$ aligns the selection order at every search step, while $\mathcal{L}_{\texttt{SelT}}$ enforces the stopping criterion at the target terminal step. We combine their sum with the classification loss according to Eq.~\ref{eq:overall_objective} and update $\theta$. Algorithm~\ref{algo} summarizes the complete fine-tuning procedure.


\subsection{Training Efficiency}\label{sec:training_efficiency}

To reduce the computational overhead of search-based attribution, we use several implementation-level optimizations.
\textit{1) Constrained search space.} We partition each image into 16 regions and restrict each candidate pool to four regions, substantially reducing the search space.
\textit{2) Sparse regularization.} We apply regularization once every 20 iterations, which empirically preserves the performance of full regularization.
\textit{3) Parallelization.} We reimplement the attribution-search and ranking-loss procedures to maximize parallel processing during the forward pass. Empirically, a batch of eight samples provides the best throughput, reducing the per-epoch training time from 11.6 hours with naive sequential processing to 5.4 hours, comparable to CGC and faster than CLIP-AFT.
Unlike CGC~\cite{cgc22}, our method avoids second-order gradients and therefore requires substantially less GPU memory.
The detailed comparison of computational resources is summarized in Table~\ref{tab:efficiency}.

\begin{table}[h]
    \centering
    \caption{Comparison of computational resources for training ViT-B on ImageNet for one epoch using a single NVIDIA A100 GPU.}
    \begin{tabular}{c|cc}
    \toprule
    Method   & Time (h) & Memory (GB) \\ \midrule
    Baseline & 2.6      & 33.7        \\
    CGC~\cite{cgc22}      & 5.4      & 64.2        \\
    CLIP-AFT~\cite{clip_aft25} & 10.2     & 43.9        \\ \rowcolor{gray!20}
    Ours & 5.4      & 38.9       \\ \bottomrule
    \end{tabular}
    \label{tab:efficiency}
\end{table}


\section{Experiment}
\subsection{Experimental Setting}
\subsubsection{Dataset}
We evaluate our method on a comprehensive collection of ImageNet-based benchmarks covering in-distribution evaluation, robustness assessment, and out-of-distribution generalization. For standard evaluation, we use ImageNet~\cite{imagenet2009} as the primary benchmark, and ImageNet-ReaL~\cite{imagenetreal2020}, and ImageNet-V2~\cite{imagenetv22019} provide additional evaluations under naturally occurring distribution variations.
For robustness evaluation, we adopt ImageNet-C~\cite{imagenetc19}, ImageNet-P~\cite{imagenetc19}, and ObjectNet~\cite{objectnet19}. 
For OOD evaluation, we use ImageNet-Sketch~\cite{imagenets19}, ImageNet-R(endition)~\cite{imagenetr2021}, Stylized ImageNet~\cite{stylizedimagenet2018}, ImageNet-A~\cite{imageneta2021}, and ImageNet-SD~\cite{imagenetsd2023}, which introduce diverse domain shifts including sketch abstraction, artistic rendition, texture bias reduction, adversarially filtered samples, and diffusion-generated images. These benchmarks collectively provide a comprehensive evaluation of the model's generalization ability across diverse visual distributions.

\subsubsection{Experiment setup}
We conduct experiments using both ViT-Base/16 and ViT-Large/16 architectures. We utilize all training and validation images within these categories. We compare the existing attribution regularization methods GC~\cite{gc21}, CGC~\cite{cgc22}, DRE~\cite{dre2023}, R2ET~\cite{r2et2024}, CLIP-AFT~\cite{clip_aft25} and EGAT~\cite{egat2026}. It should be noted that while the original GC and CGC methods utilized Grad-CAM to generated attributions for training on ResNet, we employ the more recent gradient-based method Grad-ECLIP~\cite{eclipC24} to generate attributions when reproducing the experiments on ViT models. We fine-tune the pre-trained model on the ImageNet training set for 2 epochs with each regularization method. For more training details, please refer to \textit{supplementary material} \ref{sec:traing_deail}.

\begin{table*}[t]
\caption{Top-1 classification results on in-distribution, robustness, and OOD benchmarks. Higher is better for accuracy (\%), whereas lower is better for the ImageNet-C and ImageNet-P metrics.}
\centering
\resizebox{\textwidth}{!}{
\begin{tabular}{l|ccc|ccc|ccccc}
\toprule
\multirow{2}{*}{Methods} & \multicolumn{3}{c|}{Regular}                     & \multicolumn{3}{c|}{Robustness}                                     & \multicolumn{5}{c}{Out-of-Distribution}                                            \\ \cmidrule(lr){2-12} 
                        & ImageNet-1K       & ReaL           & V2             & Corrupted (mCE$\downarrow$)  & Perturbation (mCE$\downarrow$)  & ObjectNet             & Sketch         & Rendition              & Adversarial  & Stylized           & SD             \\ \midrule
ViT-B (pretrained) & 81.44          & 86.46          & 77.37          & 40.86               & 19.51               & 36.13          & 35.43          & 33.73          & 16.00          & 28.51          & 61.33          \\
+finetune               & 82.15          &    85.54            & 78.20          & 42.03               & 19.40               & 36.55          & 29.29          & 32.70          & 19.59          &        27.78        &   61.04             \\
+$\text{GC}_{GE}$~\cite{gc21}      & 82.51          & 87.39          & 78.58          & 39.78               & 18.03               & 38.78          & 36.70          & 35.72          & 18.13          & 29.86          & 63.34          \\
+$\text{CGC}_{GE}$~\cite{cgc22}      & 82.50          & 87.42          & 78.69          & 38.98               & 17.38               & 39.12          & 35.75          & 34.73          & 19.65          & 28.82          & 63.35          \\
+$\text{DRE}_{GE}$~\cite{dre2023}      & 78.77          & 84.51          & 74.44          & 49.68               & 23.43               & 35.83          & 33.20          & 33.81          & 13.55          & 28.24          & 60.89          \\
+R2ET~\cite{r2et2024}                   & 77.18          & 83.93          & 73.37          & 47.05               & 22.51               & 32.17          & 27.44          & 32.46          & 9.69           & 24.43          & 63.27          \\
+CLIP-AFT~\cite{clip_aft25}               & 79.81          & 85.87          & 76.07          & 44.45               & 21.95               & 33.55          & 36.49          & 35.24          & 12.44          & 26.56          & 61.98          \\
+EGAT~\cite{egat2026}                   & 78.78          & 85.30          & 75.37          & 50.61               & 23.69               & 36.69          & 30.56          & 33.25          & 14.36          & 26.39          & 61.59          \\ \midrule
\rowcolor{gray!20}
Ours & \begin{tabular}{c}\textbf{82.58}\\[-1mm]{\scriptsize\color{lightblue}{(+0.10\%)}}\end{tabular} & \begin{tabular}{c}\textbf{87.52}\\[-1mm]{\scriptsize\color{lightblue}{(+0.11\%)}}\end{tabular} & \begin{tabular}{c}\textbf{78.86}\\[-1mm]{\scriptsize\color{lightblue}{(+0.22\%)}}\end{tabular} & \begin{tabular}{c}\textbf{38.75}\\[-1mm]{\scriptsize\color{lightblue}{(2.59\%$\downarrow$)}}\end{tabular} & \begin{tabular}{c}\textbf{17.36}\\[-1mm]{\scriptsize\color{lightblue}{(0.12\%$\downarrow$)}}\end{tabular} & \begin{tabular}{c}\textbf{40.87}\\[-1mm]{\scriptsize\color{lightblue}{(+4.47\%)}}\end{tabular} & \begin{tabular}{c}\textbf{36.74}\\[-1mm]{\scriptsize\color{lightblue}{(+0.11\%)}}\end{tabular} & \begin{tabular}{c}\textbf{36.29}\\[-1mm]{\scriptsize\color{lightblue}{(+1.60\%)}}\end{tabular} & \begin{tabular}{c}\textbf{21.09}\\[-1mm]{\scriptsize\color{lightblue}{(+7.33\%)}}\end{tabular} & \begin{tabular}{c}\textbf{30.08}\\[-1mm]{\scriptsize\color{lightblue}{(+4.23\%)}}\end{tabular} & \begin{tabular}{c}\textbf{63.41}\\[-1mm]{\scriptsize\color{lightblue}{(+0.09\%)}}\end{tabular}\\
\bottomrule
\end{tabular}
}
\label{tab:main}
\end{table*}

\subsubsection{Evaluation Metric}
We adopt a comprehensive set of metrics to evaluate both task performance and attribution quality. Classification performance is measured by Top-1 and Top-5 accuracy (denoted as Acc @1 and Acc @5). Attribution quality is evaluated with several metrics. Stability (denoted as Stab.) is a modified IoU that measures the overlap between attribution regions before and after image flipping, with an additional penalty term to account for excessively large attribution regions. Attribution Area (denoted as Area) quantifies the fraction of image pixels highlighted by the attribution map, reflecting the precision of explanations. We also report Insertion and Deletion AUC scores ~\cite{petsiuk2018rise,DBLP:conf/icml/WangW24b} to evaluate attribution faithfulness.
Finally, Pointing Game (PG) ~\cite{DBLP:journals/corr/abs-2306-03400} and Energy Pointing Game (EPG) ~\cite{DBLP:conf/iclr/GairolaBLS25} are adopted to evaluate the spatial alignment between attribution maps and ground-truth object locations. For more implementation details of the evaluation metrics, please refer to \textit{supplementary material} \ref{sec:traing_deail}.



\begin{figure*}[t]
    \centering
    \includegraphics[width=0.8\textwidth]{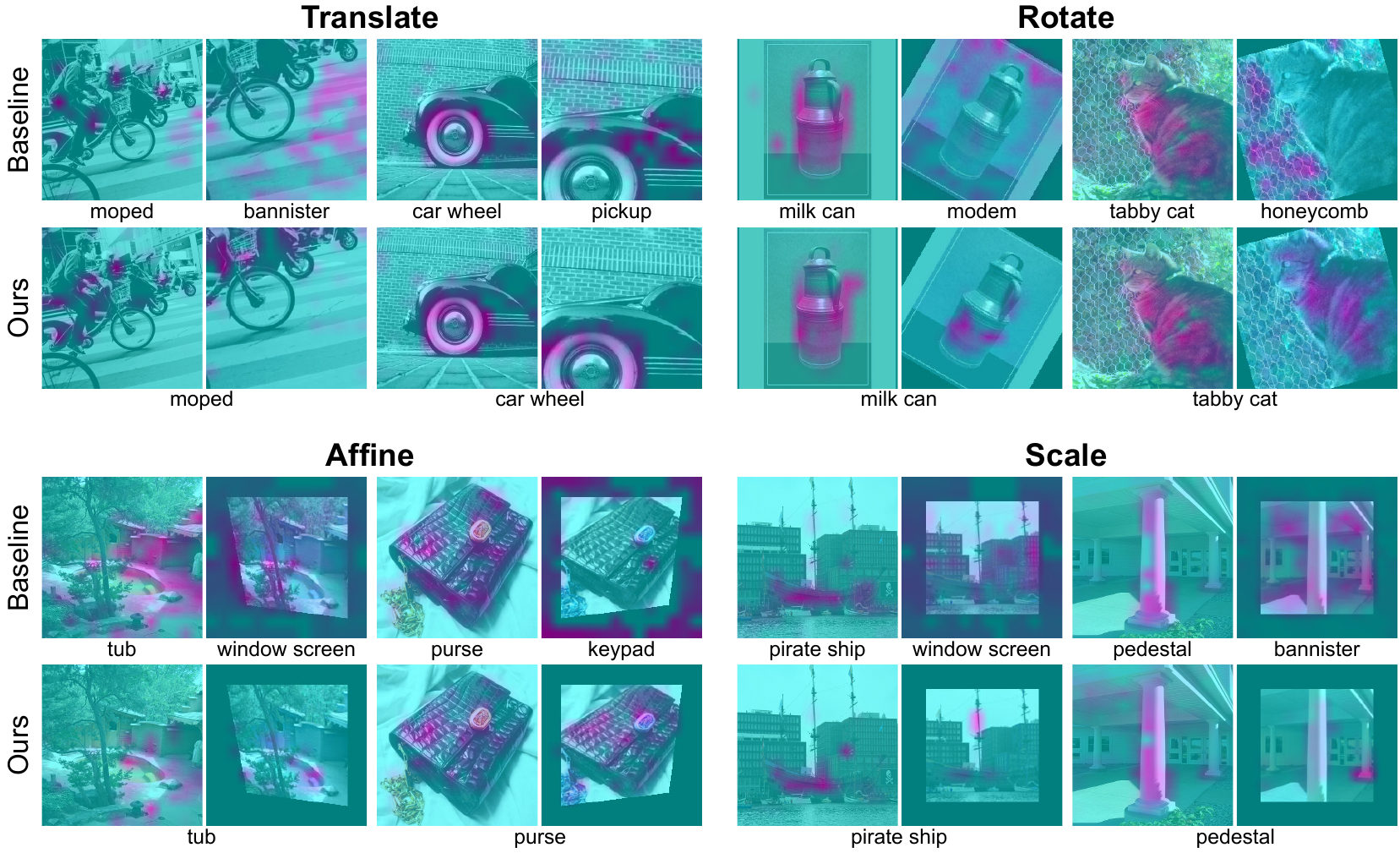} 
    \caption{Qualitative comparison of the baseline and our model under four geometric transformations. Attribution maps are generated with Grad-ECLIP on ViT-B/16, and predicted class labels are shown below each image.}
    \label{fig:robustness}
\end{figure*}

\subsection{Task Performance}

In this subsection, we evaluate in-distribution accuracy, robustness to geometric transformations and common corruptions, OOD generalization, and transfer across model architectures.

\subsubsection{In-distribution Performance Evaluation}

Table~\ref{tab:main} reports the in-distribution results. Our method obtains 82.58\%, 87.52\%, and 78.86\% Top-1 accuracy on ImageNet, ImageNet-ReaL, and ImageNet-V2, respectively. Relative to the pretrained ViT-B/16, these values correspond to gains of 1.14, 1.06, and 1.49 percentage points. Our method also exceeds the strongest non-ours result by 0.07, 0.10, and 0.17 points on the three benchmarks, showing that the regularizer preserves and slightly improves standard predictive performance.

\subsubsection{Robustness to Geometric Transformation}

We construct a transformed validation benchmark using translation (implemented by cropping), rotation, affine warping, and scaling. Every model is evaluated on the same transformed images. Transformation parameters and examples are provided in the \textit{supplementary material} (Section~\ref{sec:transformation}).


Quantitative results on ViT-B in Table~\ref{tab:main} show that our regularization method consistently outperforms baseline approaches across all transformations, confirming that the proposed attribution regularization effectively improves model robustness to transformations.
We qualitatively examine model behavior under these transformations using Grad-ECLIP~\cite{eclipC24}. We select examples that both the baseline and our model classify correctly before transformation. In Fig.~\ref{fig:robustness}, the baseline attribution can shift toward irrelevant background regions after transformation (e.g., from the motorcycle to the road in a translated image), whereas our model more consistently focuses on the object. These examples are consistent with improved evidence stability, although the quantitative support should be supplied as noted above.

\subsubsection{Robustness against Corruption}
Robustness to unexpected visual variations is crucial for deploying vision models in real-world scenarios. We therefore evaluate our method under corrupted and shifted visual conditions. As reported in Table~\ref{tab:main}, our method achieves superior robustness compared with both the vanilla ViT-B and existing adaptation methods.


Specifically, our method achieves 38.75 mCE on ImageNet-C, improving over the pretrained ViT-B/16 by 2.11 error points and over the strongest non-ours method by 0.23 points. It also achieves 17.36 on ImageNet-P, an improvement of 2.15 points over the pretrained model. The ImageNet-P metric name must be inserted after it is verified. Several comparison methods, including $\text{DRE}_{GE}$~\cite{dre2023}, R2ET~\cite{r2et2024}, and EGAT~\cite{egat2026}, reduce clean accuracy while still yielding higher robustness error; our method instead obtains both the highest ImageNet accuracy and the lowest reported robustness errors.

Furthermore, our method obtains 40.87\% accuracy on ObjectNet, exceeding the strongest baseline by 1.75 percentage points. Since ObjectNet introduces substantial variations in object pose, context, and appearance, this improvement indicates that our method encourages the model to learn more invariant and transferable visual features.

These observations confirm that the proposed attribution regularization effectively enhances the robustness of ViT against both artificial corruptions and naturally occurring distribution changes, making it more reliable for real-world visual recognition scenarios.

\subsubsection{Generalization to OOD Data}

To evaluate generalization beyond the training distribution, we test the models on five OOD benchmarks. Table~\ref{tab:main} shows that our method obtains 36.74\%, 36.29\%, 21.09\%, 30.08\%, and 63.41\% accuracy on ImageNet-Sketch, ImageNet-R, ImageNet-A, Stylized ImageNet, and ImageNet-SD, respectively. These results exceed the pretrained ViT-B/16 by 1.31, 2.56, 5.09, 1.57, and 2.08 percentage points.


Compared with existing approaches, our method also achieves the best value in every OOD column, with margins over the strongest non-ours result ranging from 0.04 points on ImageNet-Sketch to 1.44 points on ImageNet-A. Although $\text{GC}_{GE}$ and $\text{CGC}_{GE}$ improve several OOD results, neither improves consistently across all five datasets.

The superior OOD generalization performance suggests that our method encourages ViT to capture semantic features that are less dependent on domain-specific appearance cues. By improving representation robustness and reducing sensitivity to distribution variations, our method enables the model to generalize effectively to unseen visual environments.

\subsubsection{Generalization Across Different Architectures}

To evaluate transfer beyond ViTs, we apply our method to ResNet-50~\cite{he2016deep} and ConvNeXt-B~\cite{liu2022convnet}. As shown in Table~\ref{tab:architecture}, ResNet-50 improves by 0.03, 0.69, 0.63, and 0.85 percentage points under crop, rotation, affine warping, and scaling, respectively, for a mean transformed-input gain of 0.55 points; clean accuracy decreases by 0.12 points. For ConvNeXt-B, affine and scale accuracy improve by 0.79 and 0.82 points, while crop and rotation decrease by 0.13 and 0.16 points, yielding a mean transformed-input gain of 0.33 points with a 0.30-point reduction in clean accuracy. Thus, the method transfers to two convolutional backbones, although the gains are not uniform across every transformation.

\begin{table}[h]
    \centering
    \caption{ImageNet-1K Top-1 accuracy (\%) under regular inputs and geometric transformations.}
    \label{tab:rebuttal_model}
    \resizebox{0.48\textwidth}{!}{
    \begin{tabular}{c|c|ccccc}
    \toprule
    Model & Method & Regular & Crop & Rotate & Affine & Scale \\ \midrule
    \multirow{2}{*}{ResNet50} & Baseline & 80.34 & 72.58 & 67.65 & 77.82 & 75.82 \\
    & \cellcolor{gray!20}Ours & \cellcolor{gray!20}80.22 & \cellcolor{gray!20}72.61 & \cellcolor{gray!20}68.34 & \cellcolor{gray!20}78.45 & \cellcolor{gray!20}76.67 \\ \midrule
    \multirow{2}{*}{ConvNext-B} & Baseline & 83.75 & 78.07 & 75.38 & 80.04 & 78.89 \\
    & \cellcolor{gray!20}Ours & \cellcolor{gray!20}83.45 & \cellcolor{gray!20}77.94 & \cellcolor{gray!20}75.22 & \cellcolor{gray!20}80.83 & \cellcolor{gray!20}79.71 \\
    \bottomrule
    \end{tabular}
    }
    \label{tab:architecture}
\end{table}

\begin{figure*}[t]
    \centering
    \includegraphics[width=1.0\textwidth]{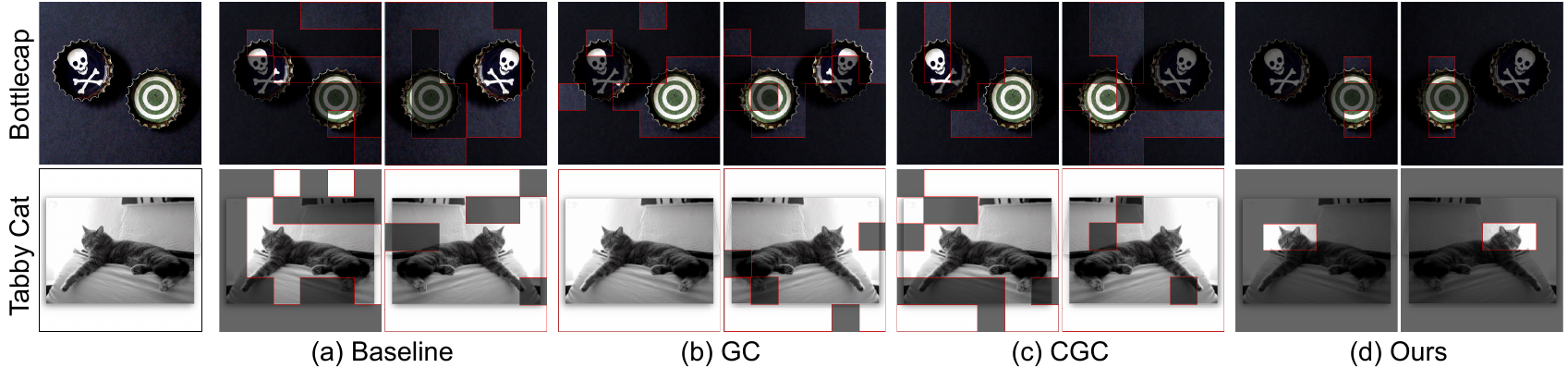} 
    \caption{Comparison of attribution results and corresponding saliency maps before and after image flipping using different methods. The model employed is ViT-Base. (Best viewed when zoomed in.)}
    \label{fig:stability1}
\end{figure*}

\begin{figure*}[t]
\centering
\includegraphics[width=1.0\textwidth]{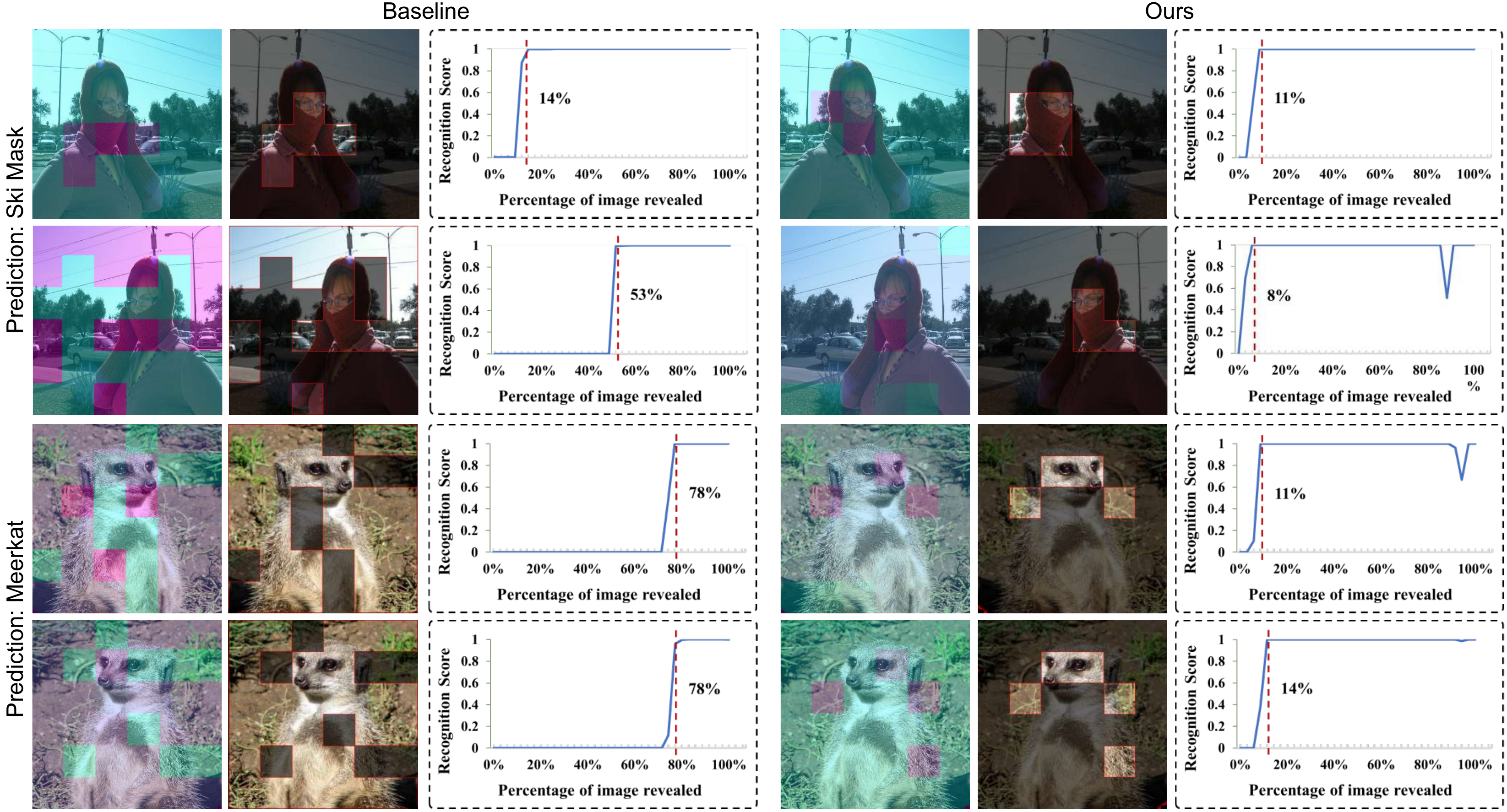} 
\caption{Visualization of attribution stability evaluation. We present attribution results, perceptual heatmaps, and Insertion curves obtained via the LIMA method, comparing model outputs before and after regularization training.}
\label{fig:attribution}
\end{figure*}

\subsection{Attribution Quality}

In this subsection, we assess whether the proposed regularization improves attribution stability, faithfulness, compactness, semantic localization, and sensitivity to learned model parameters.

\subsubsection{Stability}
Table~\ref{tab:tab1} reports ImageNet-100 classification accuracy and attribution stability. Relative to the corresponding baseline, GC and CGC reduce accuracy by 1.00--2.40 percentage points, whereas our method reduces it by only 0.28 points on ViT-B/16 and 0.27 points on ViT-L/16. Meanwhile, Stability increases from 0.14 to 0.27 on ViT-B/16 and from 0.15 to 0.23 on ViT-L/16. The qualitative comparisons in Fig.~\ref{fig:stability1} likewise show more consistent attribution regions before and after horizontal flipping. Together, these results indicate a substantial improvement in attribution stability at a small cost in classification accuracy.

\begin{table}[h]
    \centering
    \caption{ImageNet-100 classification accuracy, attribution stability, and faithfulness for ViT-B/16 and ViT-L/16.}
    \label{tab:tab1}
    \resizebox{0.48\textwidth}{!}{
    \begin{tabular}{c|l|c|c|cc}
    \toprule
    \multicolumn{1}{c|}{\small Models} & \multicolumn{1}{c|}{\small Methods} 
    & \small Accuracy ($\uparrow$) & \small Stability ($\uparrow$) 
    & \small Insertion ($\uparrow$) & \small Deletion ($\downarrow$) \\ 
    \midrule
    \multirow{4}{*}{\small ViT-B} 
    & \small Baseline & \textbf{0.9360} & 0.14 & 0.4496 & 0.1363 \\
    & \small GC~\cite{gc21} & 0.9194 & 0.14 & 0.4269 & 0.1002 \\
    & \small CGC~\cite{cgc22} & 0.9260 & 0.13 & 0.4095 & 0.3127 \\
    & \cellcolor{gray!20}\small Ours & \cellcolor{gray!20}0.9332 & \cellcolor{gray!20}\textbf{0.27} & \cellcolor{gray!20}\textbf{0.6840} & \cellcolor{gray!20}\textbf{0.0515} \\
    \midrule
    \multirow{4}{*}{\small ViT-L} 
    & \small Baseline & \textbf{0.9218} & 0.15 & 0.4308 & 0.1396 \\
    & \small GC~\cite{gc21} & 0.8978 & 0.15 & 0.3910 & 0.0855 \\
    & \small CGC~\cite{cgc22} & 0.9008 & 0.13 & 0.3858 & 0.1264 \\
    & \cellcolor{gray!20}\small Ours & \cellcolor{gray!20}0.9191 & \cellcolor{gray!20}\textbf{0.23} & \cellcolor{gray!20}\textbf{0.5494} & \cellcolor{gray!20}\textbf{0.0214} \\
    \bottomrule
    \end{tabular}
    }
\end{table}


\begin{figure*}[t]
  \centering
  \includegraphics[width=1.0\linewidth]{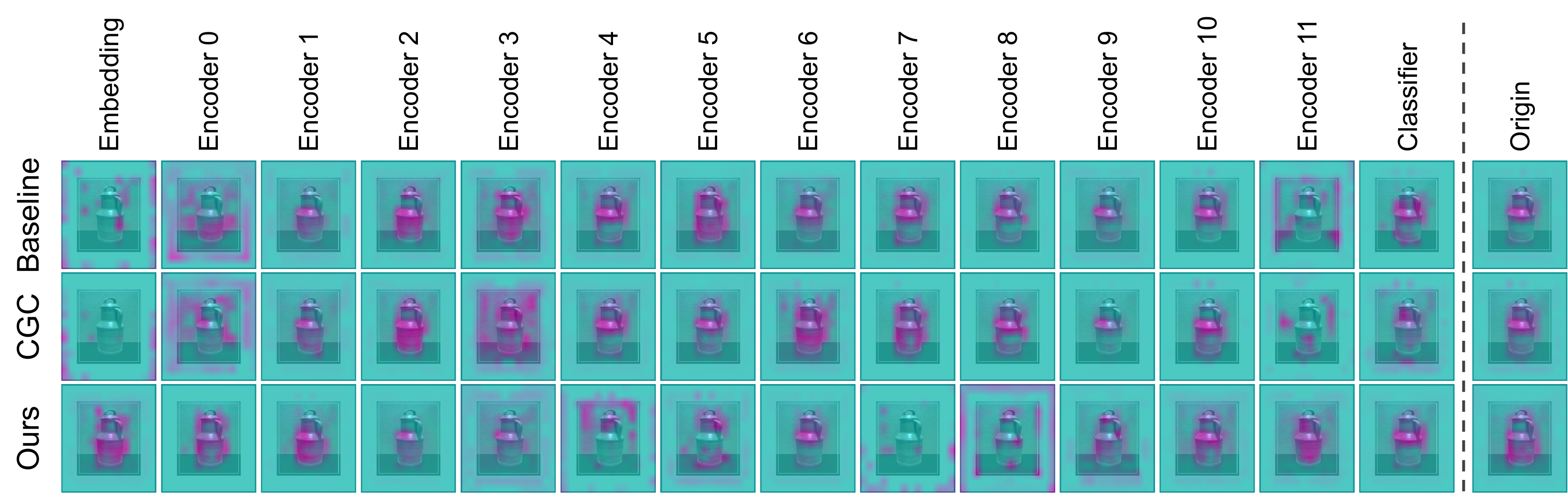}
   \caption{Sanity check heatmaps based on Grad-ECLIP.}
   \label{fig:sanity_check}
\end{figure*}


\subsubsection{Faithfulness}
We next examine attribution faithfulness using Insertion and Deletion. Table~\ref{tab:tab1} shows that our method achieves the best value for both metrics on both architectures. For ViT-B/16, Insertion increases from 0.4496 to 0.6840 and Deletion decreases from 0.1363 to 0.0515. For ViT-L/16, Insertion increases from 0.4308 to 0.5494 and Deletion decreases from 0.1396 to 0.0214.
These complementary gains indicate that highly attributed regions recover predictive evidence quickly and that removing them suppresses the prediction effectively. Fig.~\ref{fig:attribution} provides qualitative examples in which regularization produces more stable and compact search trajectories.

\subsubsection{Rationality}
We evaluate attribution compactness and semantic localization using Area, PG, and EPG in Table~\ref{tab:rationality}. LIMA attributions from our method reduce Area from 55.01\% to 31.04\% on ViT-B/16 and from 56.92\% to 37.03\% on ViT-L/16. They also improve LIMA-based PG/EPG from 0.8244/0.5508 to 0.8906/0.6829 on ViT-B/16 and from 0.8338/0.5553 to 0.8626/0.6691 on ViT-L/16.

\begin{figure}[h]
    \label{fig:pg_eclip}
    \centering
    \includegraphics[width=0.48\textwidth]{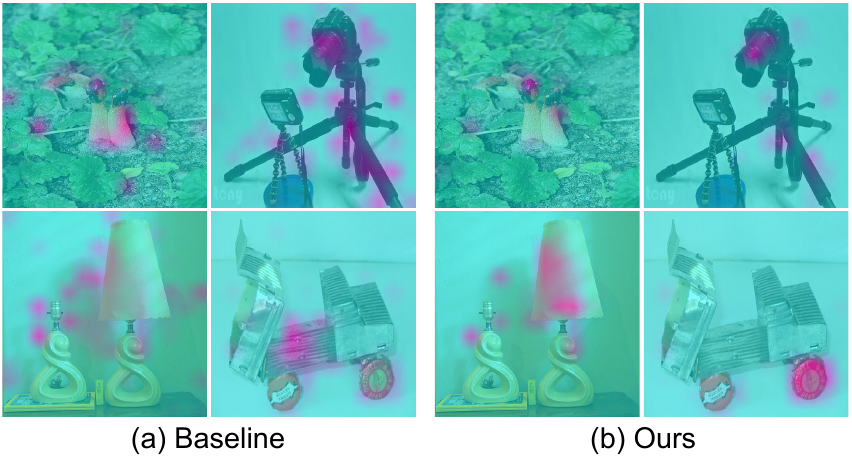} 
    \caption{Grad-ECLIP attribution maps before and after the proposed regularization.}
    \label{fig:heatmap_ge}
\end{figure}

The Grad-ECLIP localization results are more mixed. On ViT-B/16, our method produces the best EPG but not the best PG; on ViT-L/16, it is slightly below CGC on both metrics. Thus, the strongest localization gains occur for the search-based attribution used by our regularizer, while transfer to a gradient-based evaluator is partial. Figure~\ref{fig:heatmap_ge} illustrates the greater visual concentration of the regularized model's attribution.

\begin{table}[h]
    \centering
    \caption{Attribution compactness and localization on ImageNet-1K. We evaluate both the black-box attribution method LIMA and the white-box attribution method Grad-ECLIP. The best result in each column is highlighted in bold.}
    \label{tab:rationality}
    \resizebox{0.48\textwidth}{!}{
    \begin{tabular}{c|l|cccc|c}
    \toprule
    \multicolumn{1}{c|}{\small Models} & \multicolumn{1}{c|}{\small Methods} 
    & \small PG$_{\text{LIMA}}$ & \small EPG$_{\text{LIMA}}$ 
    & \small PG$_{\text{Grad-ECLIP}}$ & \small EPG$_{\text{Grad-ECLIP}}$ & \small Area ($\downarrow$) \\ 
    \midrule
    \multirow{4}{*}{\small ViT-B/16}
    & \small Baseline & 0.8244 & 0.5508 & 0.8664 & 0.7400 & 55.01\% \\
    & \small GC~\cite{gc21} & 0.8246 & 0.5442 & \textbf{0.8840} & 0.7133 & 58.99\%\\
    & \small CGC~\cite{cgc22} & 0.8278 & 0.5338 & 0.8544 & 0.6401 & 59.49\% \\
    & \cellcolor{gray!20}\small Ours & \cellcolor{gray!20}\textbf{0.8906} & \cellcolor{gray!20}\textbf{0.6829} & \cellcolor{gray!20}0.8638 & \cellcolor{gray!20}\textbf{0.7497} & \cellcolor{gray!20}\textbf{31.04\%} \\
    \midrule
    \multirow{4}{*}{\small ViT-L/16}
    & \small Baseline & 0.8338 & 0.5553 & 0.4906 & 0.4479 & 56.92\% \\
    & \small GC~\cite{gc21} & 0.8392 & 0.5406 & 0.4982 & 0.4701 & 62.47\% \\
    & \small CGC~\cite{cgc22} & 0.7990 & 0.5319 & \textbf{0.5630} & \textbf{0.5060} & 61.01\% \\
    & \cellcolor{gray!20}\small Ours & \cellcolor{gray!20}\textbf{0.8626} & \cellcolor{gray!20}\textbf{0.6691} & \cellcolor{gray!20}0.5594 & \cellcolor{gray!20}0.5053 & \cellcolor{gray!20}\textbf{37.03\%} \\
    \bottomrule
    \end{tabular}
    }
\end{table}

\subsubsection{Sanity Check}

We perform a cascading weight-randomization sanity check on ViT-B/16 following~\cite{sanity18}. As progressively more parameters are randomized, the Grad-ECLIP maps in Fig.~\ref{fig:sanity_check} change visibly, suggesting that the explanations depend on learned model parameters rather than only on parameter-independent image structure.

\subsection{Ablation Study}

In this subsection, we isolate the effects of trajectory regularization, the three attribution-objective components, and the regularization strength~$\lambda$.

\subsubsection{Constraint on Path}
The final attribution region and its search trajectory encode complementary information. To test whether intermediate selections matter, we compare direct optimization of the terminal region with the proposed trajectory-level constraint. Table~\ref{tab:ab-path} shows nearly unchanged clean accuracy (82.20\% versus 82.18\%) but improved accuracy in every reported corruption group when the path constraint is used. The largest gain occurs for Noise, from 45.46\% to 53.12\%, indicating that supervision along the selection trajectory contributes to corruption robustness.

\begin{table}[t]
\centering
\caption{Top-1 accuracy (\%) for the path-constraint ablation under clean inputs and grouped corruptions.}
\begin{tabular}{c|c|ccccc}
\toprule
\multirow{2}{*}{Method}           & \multirow{2}{*}{Regular} & \multicolumn{5}{c}{Robustness}            \\ \cmidrule(l){3-7} 
 & & Weather & Blur  & Digital & Noise & Extra \\ \midrule
w/o $\mathcal{L}_{\texttt{SelR}}$ & 82.20                    & 59.96   & 57.55 & 65.58   & 45.46 & 64.16 \\
w $\mathcal{L}_{\texttt{SelR}}$   & 82.18                    & \textbf{60.21}   & \textbf{57.64} & \textbf{65.93}   & \textbf{53.12} & \textbf{65.65} \\ \bottomrule
\end{tabular}
\label{tab:ab-path}
\end{table}

\subsubsection{Attribution Objective Function}

We ablate the confidence (Conf.), consistency (Cons.), and collaboration (Colla.) components of the LIMA objective on ViT-B/16, setting the weight of each included component to 1.0. As shown in Table~\ref{tab:ab-submodular}, the three-component objective provides the best Stability and Insertion while tying for the highest Accuracy among the regularized component configurations. No single configuration dominates every metric: Conf.+Colla. gives the smallest Area, while Conf.+Cons. gives the highest PG. The results therefore support the full objective as an overall trade-off rather than as the best configuration in every column.

\begin{table}[h]
    \centering
    \setlength{\tabcolsep}{3pt}
    \small
    \caption{Objective-component ablation on ImageNet-100 using ViT-B/16. PG is calculated with LIMA.}
    \resizebox{0.48\textwidth}{!}{
        \begin{tabular}{@{}ccc|ccccc@{}}
        \toprule
        \multicolumn{3}{c|}{\footnotesize Objective Function}                               & \multirow{2}{*}{\footnotesize Accuracy ($\uparrow$)} & \multirow{2}{*}{\footnotesize Stability ($\uparrow$)} & \multirow{2}{*}{\footnotesize Area ($\downarrow$)} & \multirow{2}{*}{\footnotesize Insertion ($\uparrow$)} & \multirow{2}{*}{\footnotesize PG ($\uparrow$)} \\ \cmidrule(r){1-3}
        \footnotesize Conf.     & \footnotesize Cons.      & \footnotesize Colla.                          &                      &                            &                            &                            &                            \\ \midrule
        \CheckmarkBold & \XSolidBrush   & \multicolumn{1}{c|}{\XSolidBrush}   &     0.9320                 &   0.25                         &                   55.02\%         &                  0.4620          &        0.8366                    \\
        \XSolidBrush   & \CheckmarkBold & \multicolumn{1}{c|}{\XSolidBrush}   &   0.9318                   &     0.09                       &              74.00\%              &       0.2778                     &            0.9124                \\
        \XSolidBrush   & \XSolidBrush   & \multicolumn{1}{c|}{\CheckmarkBold} &    0.9306                  &         0.12                   &         58.04\%                   &         0.4001                   &      0.8730                      \\
        \CheckmarkBold & \CheckmarkBold & \multicolumn{1}{c|}{\XSolidBrush}   &    0.9332                  &         0.05                   &        80.78\%                    &           0.2064                 &         \textbf{0.9220}                   \\
        \CheckmarkBold & \XSolidBrush   & \multicolumn{1}{c|}{\CheckmarkBold} &    0.9288                  &       0.26                     &         \textbf{32.85\%}                   &          0.6661                  &        0.9214                    \\
        \XSolidBrush   & \CheckmarkBold & \multicolumn{1}{c|}{\CheckmarkBold} &       0.9252               &     0.16                       &                46.79\%            &        0.4543                    &         0.8360                   \\
        \CheckmarkBold & \CheckmarkBold & \multicolumn{1}{c|}{\CheckmarkBold} &   \textbf{0.9332}                   &      \textbf{0.27}                      &      37.03\%                      & \textbf{0.6840} & 0.8906 \\ \bottomrule
        \end{tabular}
    }
    \label{tab:ab-submodular}
\end{table}

\subsubsection{Regularization Strength $\lambda$}
We systematically evaluate the impact of the attribution regularization weight ($\lambda$) on the task performance and attribution quality of the ViT-B model, as reported in Table \ref{tab: lambda}. 
When $\lambda$ increases from $0$ to $1.00$, classification accuracy (Acc) experiences only a marginal decrease from $0.9360$ to $0.9282$, confirming that the primary task performance is well-maintained.
The weight $\lambda=0.10$ appears to be a favorable balance point, where attribution quality metrics like PG ($\uparrow$) and EPG ($\uparrow$) see significant gains (e.g., PG: $0.8244 \rightarrow 0.8906$) with a negligible accuracy drop (less than $0.003$). Excessively high weights (e.g., $\lambda=1.00$) may improve stability but can introduce instability in other metrics.
The results validate that the choice of $\lambda$ provides a tunable knob to balance between task accuracy and explanation quality, with an intermediate range ($\lambda = 0.01 \thicksim 0.50$) offering the most favorable trade-off.

\begin{table}[h]
    \centering
    \setlength{\tabcolsep}{3pt}
    \caption{Regularization-strength ablation on ImageNet-100 using ViT-B/16. PG and EPG are calculated with LIMA.}
    \resizebox{0.48\textwidth}{!}{
    \begin{tabular}{c|c|c|c|c|cc}
    \toprule
     $\lambda$ & Accuracy ($\uparrow$) & Stab. ($\uparrow$) & Insertion ($\uparrow$)  & Area ($\downarrow$) & PG ($\uparrow$) & EPG ($\uparrow$) \\ \midrule
    0 & \textbf{0.9360} & 0.14                   & 0.4496                              & 55.01\%                   & 0.8244          & 0.5508           \\
    0.01                 & 0.9346           & 0.29                   & 0.7050                            & \textbf{24.31\%} & 0.8744          & 0.6788           \\
    0.10                  & 0.9332           & 0.27                   & 0.6840                          & 24.48\%                   & 0.8906          & 0.6829           \\
    0.50                  & 0.9316           &  0.27                 &       0.6409                   &        29.88\%         &     0.8840       & 0.6641            \\
    1.00                  & 0.9282           & \textbf{0.34}                   & \textbf{0.7063} & 28.99\%                   & \textbf{0.9172}   & \textbf{0.7104}  \\ \bottomrule
    \end{tabular}
    }
    \label{tab: lambda}
\end{table}

\section{Conclusion}


This paper proposed an annotation-free attribution regularization framework to encourage consistent evidence reliance under label-preserving geometric transformations. By deriving compact and class-discriminative supervision through decision-aware submodular search, the proposed framework avoids the limitations of gradient-based attribution maps, whose consistency may not necessarily correspond to the actual decision process of the model. We further introduced the Submodular Ranking Loss, which leverages selection-ranking and selection-truncation objectives to provide differentiable supervision for aligning attribution search trajectories and their stopping points. Extensive experiments across multiple datasets and architectures demonstrate that our method consistently improves attribution stability, faithfulness, compactness, and semantic alignment, while enhancing robustness to geometric transformations with only marginal degradation in clean-image accuracy. These results highlight the potential of faithful attribution as an effective source of self-supervision for improving the consistency of model decision-making, beyond merely producing visually consistent explanations.


%

\bibliographystyle{IEEEtran}
\bibliography{main}

\appendices

\renewcommand{\thesection}{A\arabic{section}}
\renewcommand{\thesubsection}{\thesection.\arabic{subsection}}
\renewcommand{\thesubsubsection}{\thesubsection.\arabic{subsubsection}}

\renewcommand*{\thefigure}{A\arabic{figure}}
\renewcommand*{\thetable}{A\arabic{table}}
\renewcommand{\theequation}{A\arabic{equation}}
\renewcommand{\thealgocf}{A\arabic{algocf}}

\section{Introduction of Submodular Selection}
\label{append:submodular}

Submodularity is a property of set functions that characterizes diminishing marginal returns. 
Let $\mathcal{V}$ be a finite ground set and let 
$\Psi:2^{\mathcal{V}}\rightarrow\mathbb{R}$ be a set function. 
For any two subsets $\mathcal{S}_a\subseteq\mathcal{S}_b\subseteq\mathcal{V}$ 
and any element $\alpha\in\mathcal{V}\setminus\mathcal{S}_b$, 
$\Psi$ is submodular if
\begin{equation}
\label{eq:monotonical}
\Psi(\mathcal{S}_a\cup\{\alpha\})-\Psi(\mathcal{S}_a)
\geq
\Psi(\mathcal{S}_b\cup\{\alpha\})-\Psi(\mathcal{S}_b).
\end{equation}
In other words, the marginal gain obtained by adding an element decreases as the selected set becomes larger. 
A set function is monotone non-decreasing if adding an element never decreases its value.

Submodular optimization is particularly suitable for selecting a compact and informative subset from a large candidate set. 
Under a cardinality constraint, maximizing a normalized, monotone non-decreasing submodular function with a greedy algorithm admits a classical $(1-1/e)$ approximation guarantee \cite{NemhauserWF78}. 
Therefore, submodular selection provides an efficient alternative to exhaustive subset enumeration, while naturally producing an ordered sequence in which elements with larger marginal contributions are selected earlier.

This property is closely related to attribution region search. 
Given a set of image subregions, our objective is to identify an ordered subset that progressively recovers the evidence supporting the model prediction. 
At the beginning of the search, a region that contains highly discriminative evidence is expected to produce a large improvement in the interpretation objective. 
As more relevant regions are selected, the remaining regions tend to provide increasingly redundant or complementary information, resulting in smaller marginal gains. 
Consequently, a greedy submodular-style search naturally prioritizes the most influential regions and assigns lower ranks to regions that mainly explain residual evidence.

Regarding the objective $\Psi$ defined in Eq.~\ref{eq:subf}, Chen et al.~\cite{lima24} established that, under the assumption that a newly selected region contributes positively to the model interpretation, the objective is monotone non-decreasing and satisfies the diminishing-return condition in Eq.~\ref{eq:monotonical}. 
Nevertheless, in practical neural networks, complex interactions among image regions may prevent an attribution objective from being exactly submodular for every possible subset. 
Our application therefore does not require the objective to exhibit perfect global submodularity in all cases. 
Instead, it exploits the approximate diminishing-return structure observed during region selection. 
Even when the theoretical conditions are only approximately satisfied, the greedy procedure remains effective for ordered attribution: it selects regions with the largest explanatory gains first and then progressively incorporates regions that provide additional, non-redundant evidence.

\section{Experiment Setting}
\label{sec:traing_deail}
\subsection{Training Details}
During the model fine-tuning phase, we set the initial learning rate to 2e-4 and applied data augmentation including random horizontal flipping and random cropping. For GC and CGC experiments, the initial learning rate was set to 2e-4 with regularization weights $\lambda=0.5$ and 0.1 respectively. For our method, the initial learning rate was with 5e-5, while the weights for Confidence, Consistency, and Collaboration scores were all set to 1. All experiments used the AdamW optimizer with Cosine Annealing learning rate schedule.

\subsection{Evaluation Implementation}
We provide a detailed description of the implementation for evaluation using LIMA. Our evaluation methodology references \cite{vps25}.

Insertion and Deletion AUC scores: We first employ LIMA to progressively search from a completely black image to the full image, recording the consistency score and collaboration score throughout the process. The AUC of the curve formed by plotting the consistency score against the search proportion yields the Insertion AUC. Similarly, the AUC of the curve formed by (1 - collaboration score) versus the search proportion gives the Deletion AUC.

Stability: We first use LIMA to obtain the minimal decision regions for both the original and flipped images, then calculate their Intersection over Union (IoU). When the area of the union of the decision regions before and after flipping exceeds two-thirds of the total image area, the IoU value is multiplied by a penalty coefficient of $(1 - Area_{union} / Area_{image})$.

Point Game: For a given sample, we first use LIMA to obtain the smallest image region that supports a decision confidence exceeding 0.8. The sample receives a score of 1 if this region overlaps with the manually annotated bounding box, otherwise it receives a score of 0.

Energy Point Game: We similarly begin by using LIMA to obtain the minimal decision region. If this region overlaps with the labeled target bounding box, the sample's score is calculated as the area of overlap divided by the total area of the decision region. If there is no overlap, the score remains 0.

\section{Examples of Image Transformation}
\label{sec:transformation}
\begin{figure}[h]
  \centering
  \begin{tabular}{@{}c@{}}
    \includegraphics[width=1.0\linewidth]{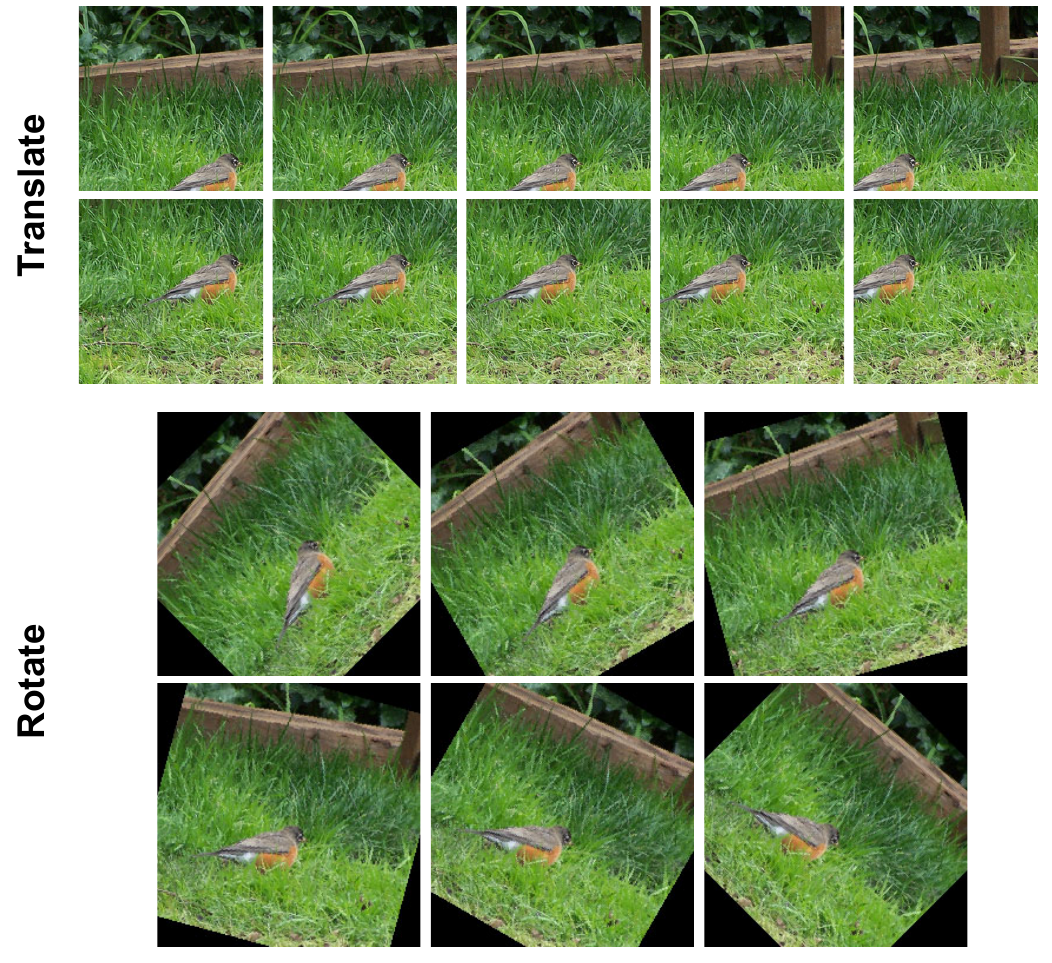}\\[-1pt]
    \includegraphics[width=1.0\linewidth]{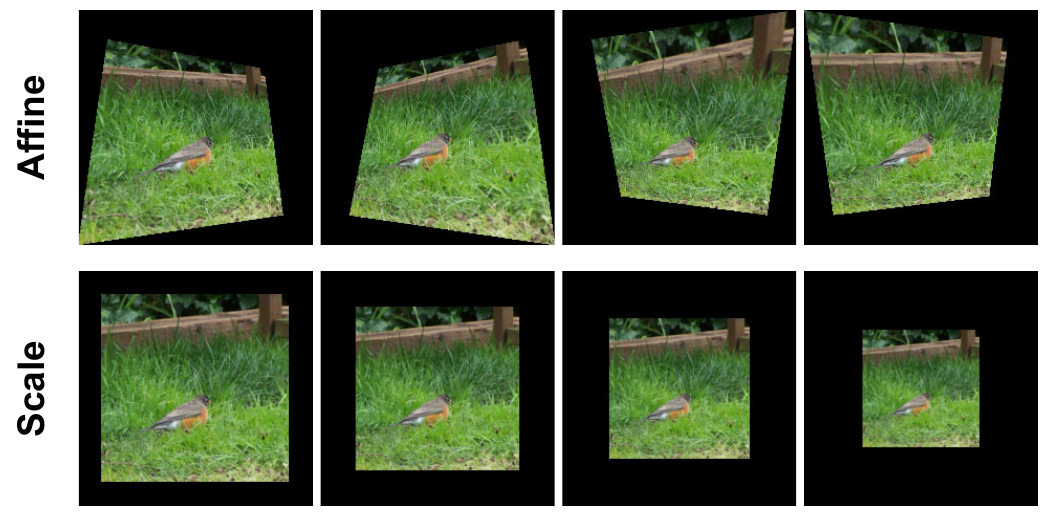}
  \end{tabular}
  \caption{Examples of enhanced image transformation.}
  \label{fig:transform}
\end{figure}

For translation, images are first resized along the shorter side to 336 pixels, then uniformly cropped into ten 224×224 patches.
The other three image transformations are applied after first resizing images along the shorter side to 256 pixels and performing center-cropping to 224×224 patches. This approach effectively simulates the translation effect of the target object within the image.
For rotation, we rotate the entire image by six angles: {15°, 30°, 45°, 315°, 330°, 345°}.
For affine, affine transformations with elastic stretching are applied to the four corners of the images. The transformation starts from the four corners of the image, which are then stretched to the following terminal points respectively: \{(56, 56), (195, 28), (223, 223), (28, 195)\}, \{(28, 28), (177, 56), (195, 195), (0, 223)\}, \{(0, 0), (195, 28), (177, 177), (28, 195)\}, \{(28, 28), (223, 0), (195, 195), (56, 177)\}. This simulates four distinct observational perspectives.
For scaling, each image is first resized to 224×224, then rescaled using factors of {0.5, 0.6, 0.7, 0.8}.
Figure \ref{fig:transform} illustrates examples of the image transformations.

\section{Visualization Results of Robustness.}
\begin{figure*}[t]
  \centering
  \begin{tabular}{@{}c@{}}
    \includegraphics[width=1.0\linewidth]{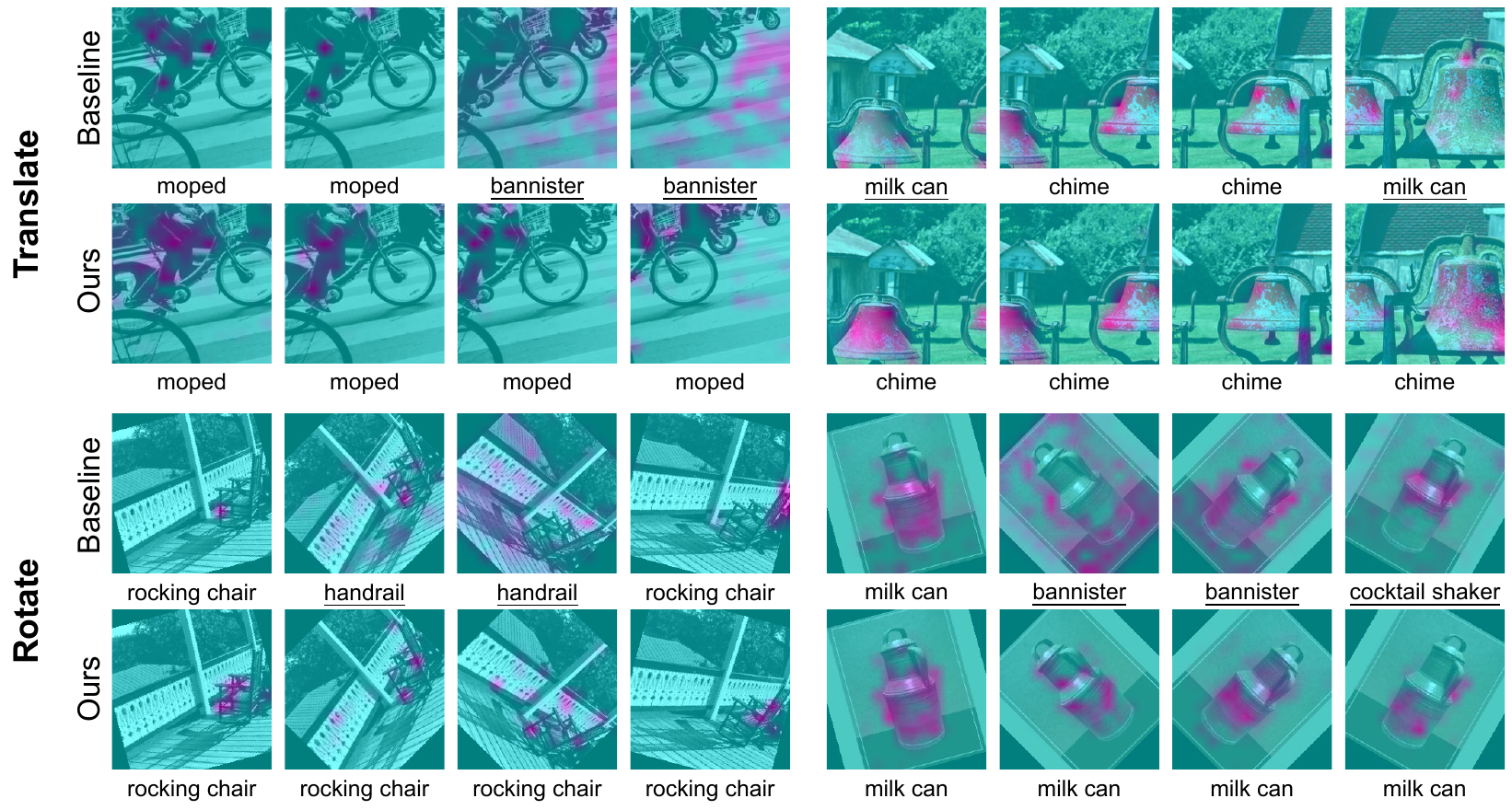}\\
    \includegraphics[width=1.0\linewidth]{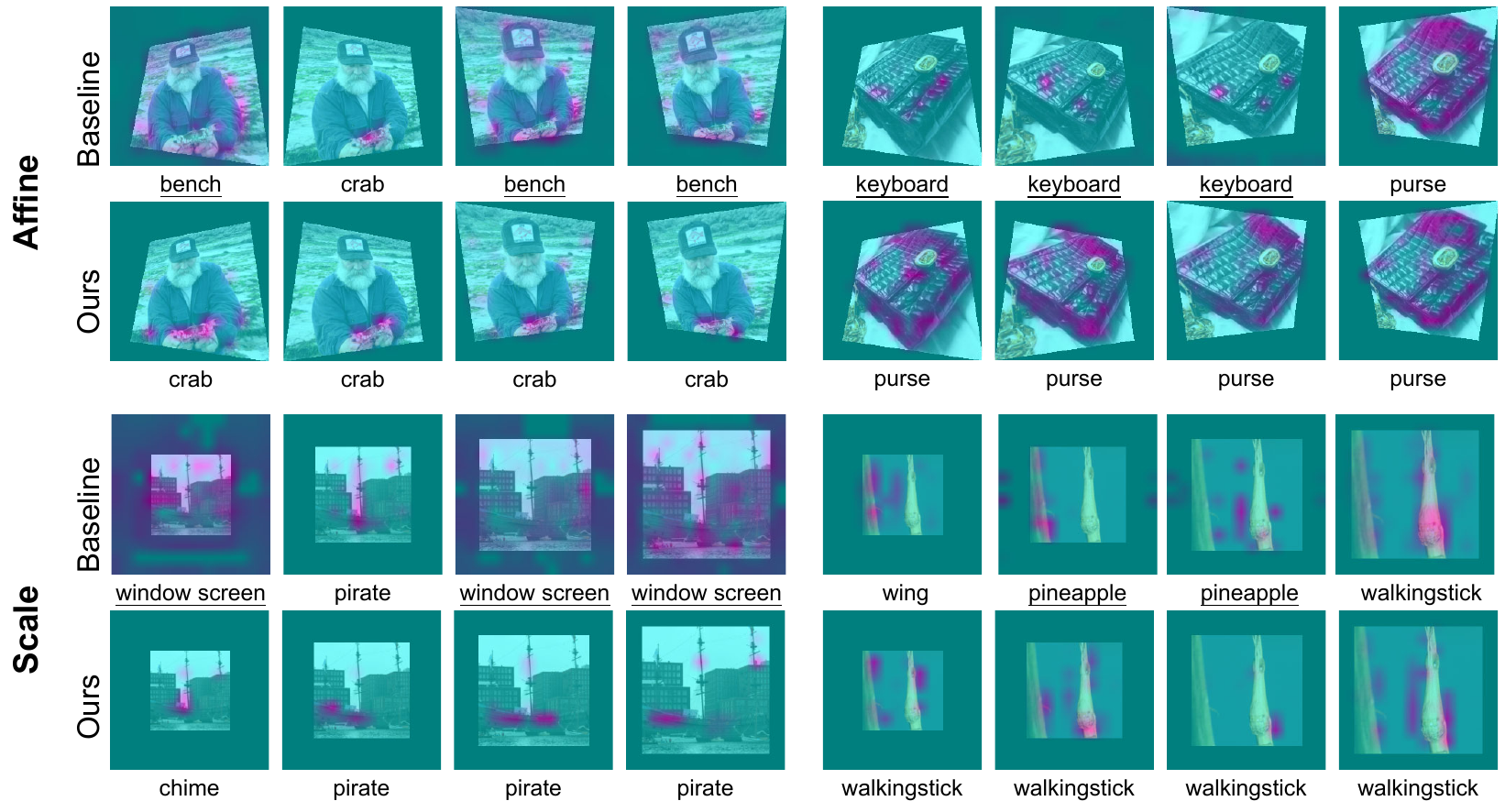}
  \end{tabular}
  \caption{Visualization of robustness evaluation. The attribution heatmaps are generated by Grad-ECLIP. The predicted labels are indicated below the corresponding images, with incorrect predictions \underline{underlined}.}
  \label{fig:robust_example}
\end{figure*}

We provide additional visualization results by including more transformed samples along with their corresponding model predictions and attribution maps, as illustrated in the \ref{fig:robust_example}. Our analysis reveals that our regularization method enhances model robustness primarily in two aspects: First, it significantly improves the model's perceptual robustness to key regions under various image transformations, effectively preventing the model from shifting its attention to irrelevant background areas or other objects. Second, even when the model's focus remains unchanged after transformation, our approach enhances its discriminative robustness to transformed features. For instance, the baseline model misclassifies a rotated milk can as a cocktail shaker, whereas our method maintains correct identification.

\end{document}